# A 3DGS-Driven Dynamic Viewpoint and Vibrotactile Framework for Subsea Teleoperation Validated via fNIRS

Fang Xu, Ph.D.[1]; Tianyu Zhou, Ph.D.[2]; Ruitong Tian[3];
Md Jahidul Islam, Ph.D.[4]; Jing Du Ph.D.[5*]

*Abstract*—**Teleoperating Remotely Operated Vehicles (ROVs) in flooded, highly cluttered infrastructure presents a severe perceptual bottleneck due to narrow 2D egocentric views and subsea communication latency. We present an integrated multimodal teleoperation architecture built on a ROS-Unity framework that decouples proactive spatial planning from reactive boundary avoidance. It replaces static visual feeds with a custom Dynamic Adaptive Viewpoint System (DAVS). DAVS utilizes continuous optimization and real-time 3D Gaussian Splatting (3DGS) to autonomously synthesize an occlusion-free, exocentric perspective from onboard state estimation. To distribute sensory load, this visualization coupled with a torso-mounted vibrotactile suit that maps local obstacle clearance to reactive haptic proximity cues. We validated the efficacy of this multimodal architecture through a controlled human-subject study (N=30) navigating a BlueROV2 in a complex, simulated underwater facility. Using a 3×4 repeated-measures design, this study evaluated three interaction modalities (Egocentric, Haptic, Exocentric) across four communication delays (0.0 to 1.0 s). System performance was objectively quantified using functional near-infrared spectroscopy (fNIRS) to capture task-evoked hemodynamic responses in prefrontal regions, alongside traditional behavioral metrics. Results demonstrate that while reactive haptics improve path adherence at minimal delay, the 3DGS-driven exocentric vision provides critical latency resilience, outperforming other modalities at severe delays (0.5-1.0 s). Crucially, fNIRS revealed a cognitive disengagement phenomenon: at critical delays, standard egocentric teleoperation overwhelmed working memory, collapsing prefrontal activation. Conversely, the proactive 3DGS-DAVS context sustained executive control, proving spatially grounded assistance optimizes operator endurance in degraded teleoperation.**

## I. Introduction

Remotely operated vehicles (ROVs) remain a practical choice for inspection and intervention in underwater environments where direct human access is unsafe, costly, or infeasible. In advanced deployment scenarios such as navigating flooded industrial facilities, subsea manifolds, and complex ship structures, operators are forced to maneuver through highly cluttered spaces where visibility is limited, and free-space margins are remarkably small. While recent advances in pure autonomy show promise, low-texture environments, turbidity, and complex 3D geometry routinely degrade state estimation, necessitating human-in-the-loop teleoperation to ensure mission success [1, 2].

However, navigating these environments amplifies the difficulty of teleoperation to the point of a severe perceptual bottleneck. The operator must continuously infer spatial clearance and complex global geometry from a partial, 2D egocentric camera view. This forces the human brain to constantly perform heavy mental rotation and spatial updating to anticipate clearance and regulate speed near obstacles. Even when the robotic platform is mechanically capable, mission success often hinges on the human's ability to maintain precise control under this immense cognitive load. Furthermore, this spatial inference burden is frequently exacerbated by severe communication latency inherent to acoustic subsea data links, which creates a critical temporal disconnect between operator input and visual feedback [3, 4]. To reduce operator burden, prior work has often focused on augmenting the primary egocentric camera feed with additional visual context. Exocentric viewpoints, reconstructed local geometry, and mixed-reality overlays can reduce the need for mental rotation, especially in tight turns and narrow passages [5, 6]. Yet, relying solely on static visual augmentation still channels all spatial information through a single sensory modality, and fixed third-person cameras frequently suffer from severe structural occlusion in highly dense environments [7-9]. Furthermore, teleoperation interfaces are rarely evaluated with objective physiological metrics to determine if these aids reduce cognitive demand or merely shift it to different mental processes [10].

This paper introduces a unified ROV teleoperation system designed to resolve this perceptual bottleneck through multimodal cognitive offloading. We combine two complementary components aimed at reducing cognitive demand without overwhelming the operator's visual processing: with a custom 3DGS-DAVS that autonomously synthesizes a photorealistic, occlusion-free third-person perspective derived from real-time state estimation, and haptic proximity cues that convey local clearance information through intuitive, torso-mounted tactile signals. We posit that this dynamic visual augmentation acts as a proactive spatial planning aid, while the haptic feedback serves as a reactive local safety shield. Crucially, we validate this system through a controlled 3×4 repeated-measures human-subject study in a highly simulated, cluttered underwater facility, measuring

Fang Xu, Postdoctoral Associate, Informatics, Cobots and Intelligent Construction (ICIC) lab, Dept. of Civil and Costal Engineering, Weil Hall 360, University of Florida; Email: xufang@ufl.edu

Tianyu Zhou, Postdoctoral Associate, Informatics, Cobots and Intelligent Construction (ICIC) lab, Dept. of Civil and Costal Engineering, Weil Hall 360, University of Florida; Email: zhoutianyu@ufl.edu

Ruitong Tian, Ph.D. Student, Informatics, Cobots and Intelligent Construction (ICIC) lab, Dept. of Civil and Costal Engineering, Weil Hall 360, University of Florida; Email: ruitongtian@ufl.edu

Md Jahidul Islam, Assistant Professor, RoboPI Lab, Dept. of Electrical and Computer Engineering, University of Florida; jahid@ece.ufl.edu

Jing Du, *Professor, Informatics, Cobots and Intelligent Construction (ICIC) Lab, Dept. of Civil and Costal Engineering, 460F Weil Hall, University of Florida, Gainesville, FL 32611 (corresponding author); Email: eric.du@essie.ufl.edu

both task performance and neurophysiological operator state under varying network delays. To quantify cognitive workload beyond subjective self-reports, we integrate fNIRS to analyze task-evoked hemodynamic responses in prefrontal regions associated with executive control and workload regulation. This enables us to evaluate a core question directly relevant to deployed teleoperation: when spatial awareness is distributed across dynamic visual and somatosensory channels under severe latency, does it fundamentally preserve the operator's executive function and task execution. To explicitly separate our technical novelty from prior arts, the proposed DAVS continuously optimizes the camera viewpoint by minimizing a visibility-occlusion objective defined over the active MonoGS scene representation, while strictly maintaining temporal smoothness constraints. Unlike static third-person cameras or pre-defined viewpoints, DAVS adapts the perspective online using real-time robot pose estimates and predicted obstacle proximity. This ensures a persistent, optimal line-of-sight without relying on offline environmental pre-training.

This paper makes three main contributions: *(a)*: A 3DGS-Driven Multimodal Architecture: We present a custom ROS-Unity framework for confined ROV navigation that seamlessly couples a novel Dynamic Adaptive Viewpoint System and real-time 3D Gaussian Splatting to prevent environmental occlusion, integrated alongside reactive haptic proximity cues in a single operator interface.

*(b)*: Evaluating Latency Resilience: Through a 3×4 factorial study design, we isolate and evaluate the distinct navigational impacts of proactive visual spatial offloading versus reactive somatosensory boundary avoidance under extreme simulated subsea communication delays (up to 1.0 s).

*(c)*: Neurophysiological Discovery of Cognitive Disengagement: We provide objective fNIRS-derived validation demonstrating that while standard egocentric interfaces induce working memory collapse and cognitive disengagement at critical delays, 3DGS-driven exocentric offloading successfully sustains prefrontal executive control and preserves operator performance.

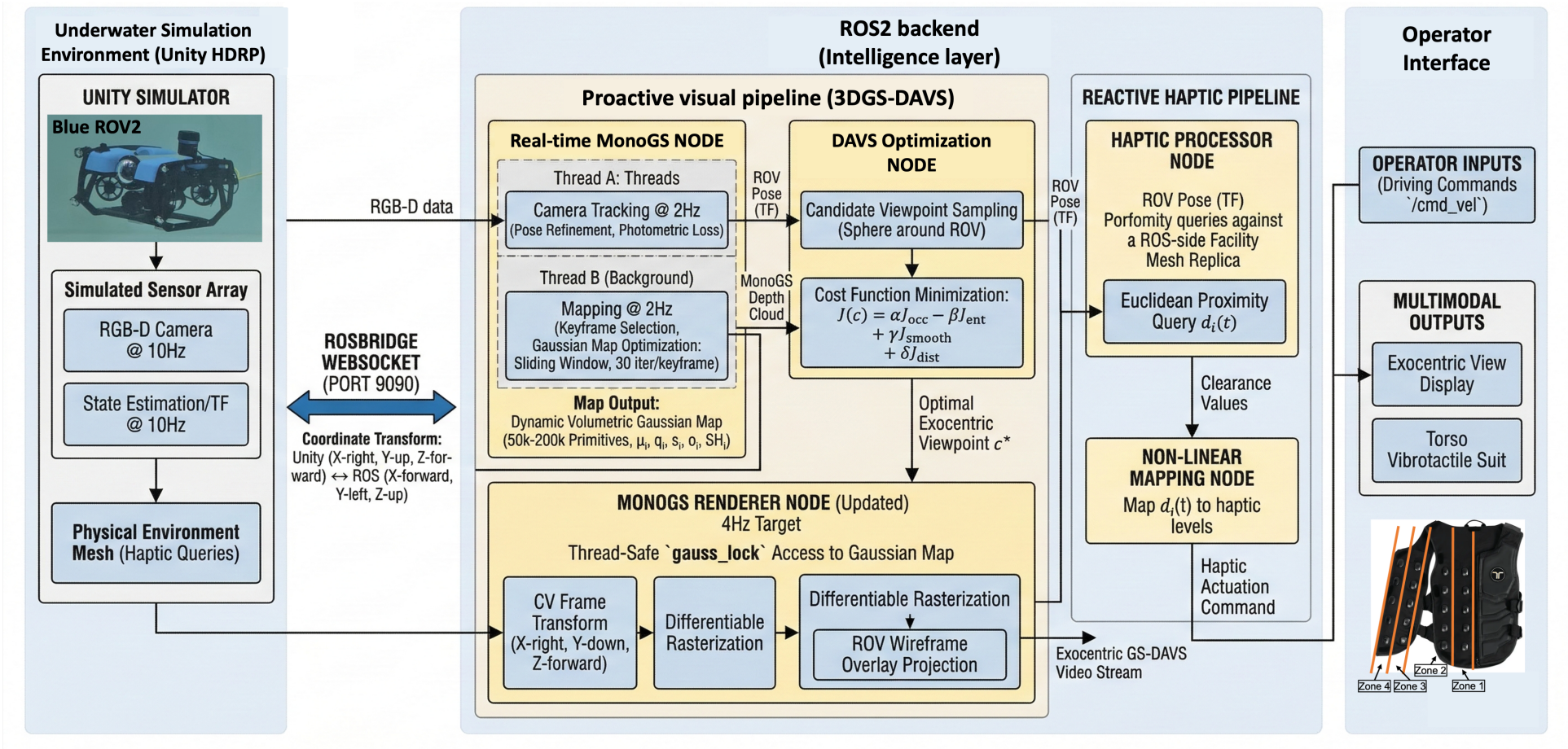

Figure 1. System architecture of the multimodal teleoperation framework. Proactive visual offloading (top) uses MonoGS for real-time scene mapping, allowing DAVS to dynamically optimize an occlusion-free exocentric viewpoint. Concurrently, reactive boundary avoidance queries vehicle collision geometry to drive torso-mounted haptic proximity cues.

## II. Related Work

### A. Exocentric and XR Interfaces for Teleoperation

A persistent limitation of direct teleoperation is that operators must infer global spatial structure from a narrow egocentric camera stream, which increases mental transformation demands [11]. To reduce this burden, prior systems have combined camera feeds with alternative viewpoints. Early work demonstrated that coupling 3D reconstruction with projection methods improves operator understanding. Recent mixed reality interfaces broaden this by integrating depth sensing to support spatial judgment. However, in underwater domains, real-time dense mapping is severely hampered by low-texture environments and turbidity. Consequently, static exocentric viewpoints often suffer from structural occlusion. Recently, 3D Gaussian Splatting (3DGS) has emerged as a powerful alternative [12], modeling environments as 3D Gaussian primitives rasterized via differentiable alpha-compositing. While 3DGS enables photorealistic novel view synthesis, applying continuous dynamic viewpoint optimization over a 3DGS representation to provide occlusion-free, latency-resilient ROV teleoperation remains largely unexplored. Most evaluations also still focus on usability outcomes without explicitly quantifying how dynamic viewpoint augmentation changes cognitive demand.

### B. Multimodal and Haptic Feedback in Teleoperation

Because complex teleoperation heavily taxes the operator's visual processing center, relying solely on visual augmentation risks creating a sensory bottleneck. To distribute cognitive load, researchers have increasingly explored multimodal feedback, particularly incorporating the somatosensory system. Recent 2025 frameworks (such as the SubSense architecture [13]) have investigated the integration of haptic feedback and VR to enhance operator situational awareness in complex subsea environments. Vibrotactile haptic cues have been effectively used to convey collision warnings [14], proximity to obstacles, and spatial guidance without requiring the operator to avert their gaze from the primary visual interface. While visual exocentric views are highly effective for proactive macro-navigation, they can lack the visceral warning needed when a robot unexpectedly drifts toward an unseen obstacle. Conversely, while haptics provides excellent reactive clearance alerts, they cannot provide the global geometry required to plan a route. Despite this natural complementarity, teleoperation interfaces rarely evaluate the interaction effects of combining proactive dynamic visual offloading with reactive haptic offloading, leaving a gap in understanding how these modalities synergize to reduce overall operator burden.

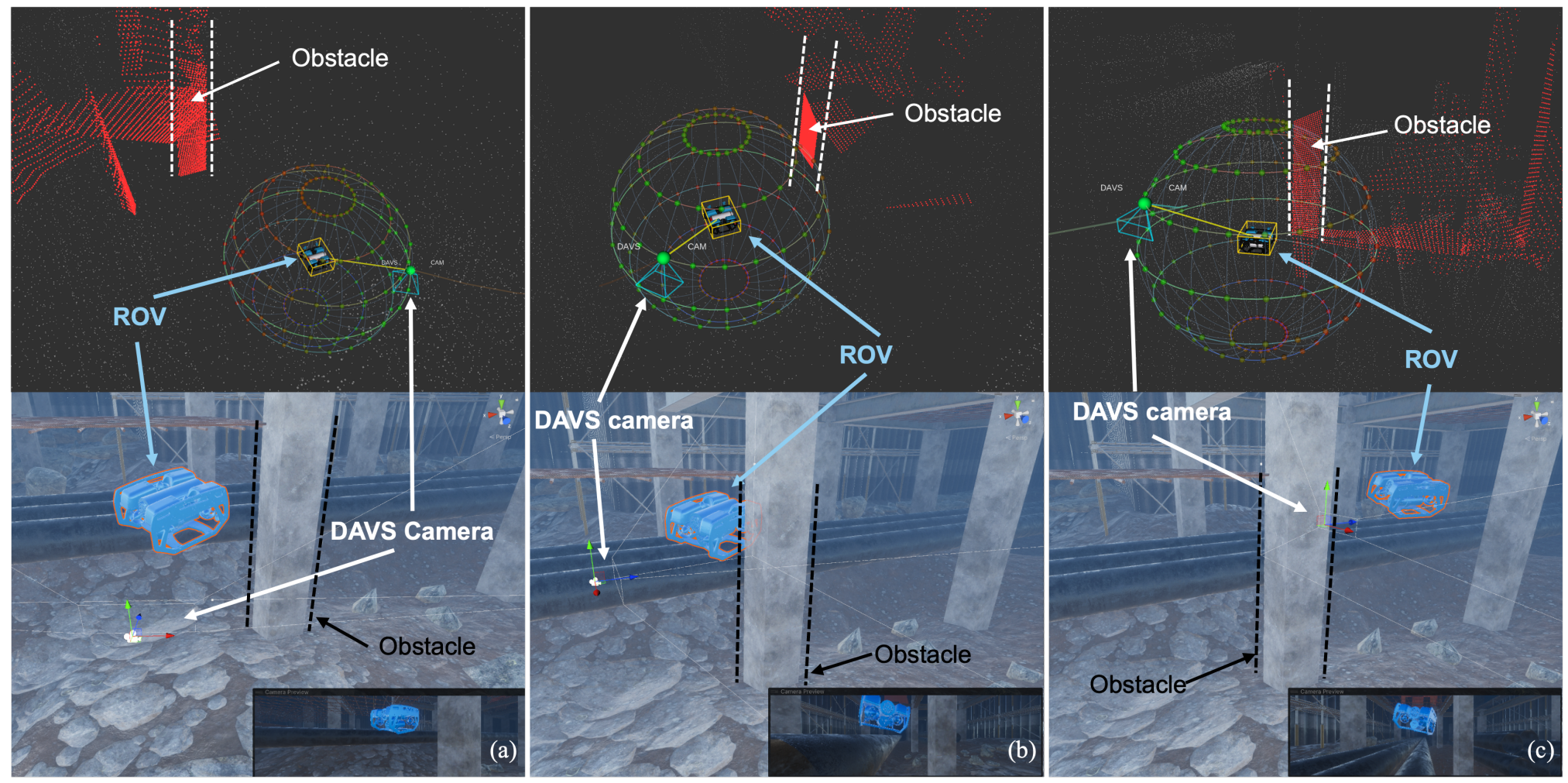

Figure 2. This RVIZ screenshot demonstrates how the proposed 3DGS-DAVS pipeline overcomes the occlusions by dynamically synthesizing an optimal, occlusion-free exocentric viewpoint.

### C. *Cognitive Workload Measurement and fNIRS in Teleoperation*

Workload in teleoperation has traditionally been assessed with subjective measures such as NASA-TLX and behavioral metrics like completion time and collisions. However, behavioral metrics alone can be insufficient to characterize operator state, since operators may maintain performance by expending higher cognitive effort. Functional near-infrared spectroscopy (fNIRS) is particularly relevant because it enables noninvasive measurement of hemodynamic responses associated with cognitive control and workload regulation in prefrontal regions [3]. Recent studies have successfully utilized fNIRS for the neural profiling of operators during complex teleoperation [15], isolating how prefrontal cortex activity correlates with task strain. However, an often-overlooked neuroergonomic metric is the "cognitive disengagement" phenomenon, where excessive task demands, such as extreme communication latency, overwhelm working memory, causing a sudden drop in prefrontal recruitment. This paper builds on these foundations by uniquely validating the combined impact of 3DGS-DAVS and haptics on objective fNIRS workload markers. We explicitly evaluate how spatial grounding prevents executive function collapse and cognitive disengagement under high-latency teleoperation conditions.

## III. System Design

### A. *System Overview*

The proposed multimodal teleoperation stack supplementing standard egocentric viewing from the conventional ROV camera with an in-house, cognitively-offloaded architecture. The system integrates real-time Monocular Gaussian Splatting SLAM (MonoGS) [16] with a Dynamic Adaptive Viewpoint System to synthesize occlusion-free, third-person camera viewpoints on the fly, eliminating the need for offline environmental pre-training.

The architecture operates through two concurrent, real-time processes. First, the MonoGS module ingests the onboard RGB stream, utilizing a dual-thread tracking and mapping pipeline to continuously estimate the ROV's 6-DoF pose while dynamically reconstructing the surrounding environment as a volumetric 3D Gaussian Splatting (3DGS) map. Concurrently, the DAVS optimizer evaluates candidate viewpoints on a sphere surrounding the vehicle, selecting the specific spatial pose that minimizes occlusion while maximizing scene information. The system then queries the active 3DGS mapping thread to render the scene from this optimal exocentric viewpoint at 10 Hz. To distribute sensory load, this proactive visual rendering is seamlessly coupled with a torso-mounted vibrotactile suit providing reactive haptic proximity cues. The full pipeline executes on ROS 2 and communicates with the BlueROV2 platform via a rosbridge websocket.

Unlike prior teleoperation interfaces that rely on static third-person views or manually controlled cameras, our system generates algorithmically optimized viewpoints using a continuously updated MonoGS scene representation. This ensures the operator maintains a persistent, occlusion-free perspective regardless of the subsea environmental complexity.

### B. *DAVS Optimization*

To synthesize the exocentric perspective, we define a virtual camera frame $V$. During Phase 3, DAVS continuously computes the optimal camera position $c^*$ from a discrete set of candidates $C$ distributed on a sphere of nominal radius $R = 1.7$ m centered on the ROV position $p_{rov}$. The optimal position $c^*$ is determined by minimizing a multi-objective cost function balancing four criteria:

$$J(c) = \alpha J_{occ}(c) - \beta J_{ent}(c) + \gamma J_{smooth}(c) + \delta J_{dist}(c)$$

where the tunable weights are set to $\alpha = 1.0, \beta = 0.3, \gamma = 0.8$, and $\delta = 0.5$.

Occlusion Cost ($J_{occ}$): To penalize structural occlusion, we evaluate the line of sight from the candidate $c$ to the ROV against the point cloud. For each point $p$, a ray-cylinder intersection test is performed along the ray direction $d = (p_{rov} - c)/|p_{rov} - c|$. A point is considered occluding if its projection $t = (p - c) \cdot d$ and perpendicular distance $r = |p - c - td|$ satisfy $0 \leq t \leq |p_{rov} - c|$ and $r < r_{cyl}$ (where $r_{cyl}$ =0.3 m). The cost $J_{occ}$ is the ratio of occluding points to total points.

Entropy Reward ($J_{ent}$): To maximize spatial awareness, points within a 60° half-angle field-of-view cone are binned by depth into $N_{bins}$=8 bins. The Shannon entropy is computed as $H = -\sum_{i=1}^{N_{bins}} p_i \log(p_i)$. The normalized reward $J_{ent} = H/\log(N_{bins})$ actively drives the camera toward information-rich viewpoints with high depth variance.

Smoothness and Distance Penalties: To enforce temporal stability, large positional jumps are penalized via $J_{smooth} = |c - c_{t-1}|^2/R^2$. Deviation from the nominal viewing sphere is penalized via $J_{dist} = \left((|c - p_{rov}| - R)/R\right)^2$.

To prevent high-frequency jitter, maintaining temporal coherence and kinematic alignment, the optimal position is filtered using an exponential moving average:

$$c_{final}(t) = (1 - \alpha_s)c_{final}(t - 1) + \alpha_s c^*(t), \text{ with } \alpha_s = 0.3$$

The orientation is strictly kinematically constrained to target the ROV. The look-at rotation matrix $R_{cam}$ is constructed using the normalized forward vector $(p_{rov} - c)$, the left vector (world up × forward), and the true up vector (forward × left), avoiding singularities when the camera is directly above or below the vehicle. This combined position and orientation yields the final rigid body transformation $T_{wv} \in SE(3)$ defining the virtual camera pose.

### *C. Monocular Gaussian Splatting SLAM and Novel View Synthesis*

Because no physical camera exists at the exocentric viewpoint computed by the DAVS module, the scene must be dynamically reconstructed from the ROV's onboard sensor data to enable real-time, photorealistic rendering. To achieve this without relying on offline environmental pre-training, we adopt a real-time MonoGS architecture [16]. The environment is represented as a volumetric set of anisotropic 3D Gaussian primitives, rasterized via differentiable alpha-compositing.

Each Gaussian primitive $i$ is parameterized by its 3D mean position $\mu_i \in R^3$ orientation quaternion $q_i \in R^4$, log-scale $s_i \in R^3$, opacity logit $o_i \in R$, and degree-3 spherical harmonic (SH) coefficients. To ensure the covariance matrix $\Sigma_i$ remains positive semi-definite during continuous optimization, the log-scale is exponentiated at render time:

$$\Sigma_i = R(q_i), diag(\exp(s_i))^2 R(q_i)^T$$

Where $R(q_i)$ is the rotation matrix derived from the quaternion. The opacity is constrained via a sigmoid activation function $\alpha_i = sig(o_i)$. The 48 SH coefficients per primitive yield view-dependent appearance modeling, which is critical for accurately representing specular reflections on the ROV hull and water-surface effects.

The tracking thread is responsible for estimating the ROV's 6-DoF camera pose in real time as new RGB frames are ingested at 10 Hz. Pose estimation is initialized using a constant-velocity motion model, providing a robust prior for the vehicle's trajectory. The exact camera pose is then refined by minimizing a photometric loss between the incoming live frame and the synthesized image rendered from the current active Gaussian map. Operating concurrently, the mapping thread continuously expands and refines the 3DGS scene representation. As the ROV explores the subsea facility, keyframes are actively selected based on spatial and angular thresholds. When a new keyframe is registered, the mapping thread inserts new Gaussian primitives into the map. Crucially, to constrain the geometry in textureless, highly turbid underwater environments where photometric gradients are weak, the optimization process incorporates monocular depth estimations. The mapping thread continuously optimizes the parameters ($\mu$,$q$,$s$,$\sigma$,$SH$) of the active Gaussians, aggressively culling primitives with low opacity or excessive scale to maintain real-time performance within strict computational bounds.

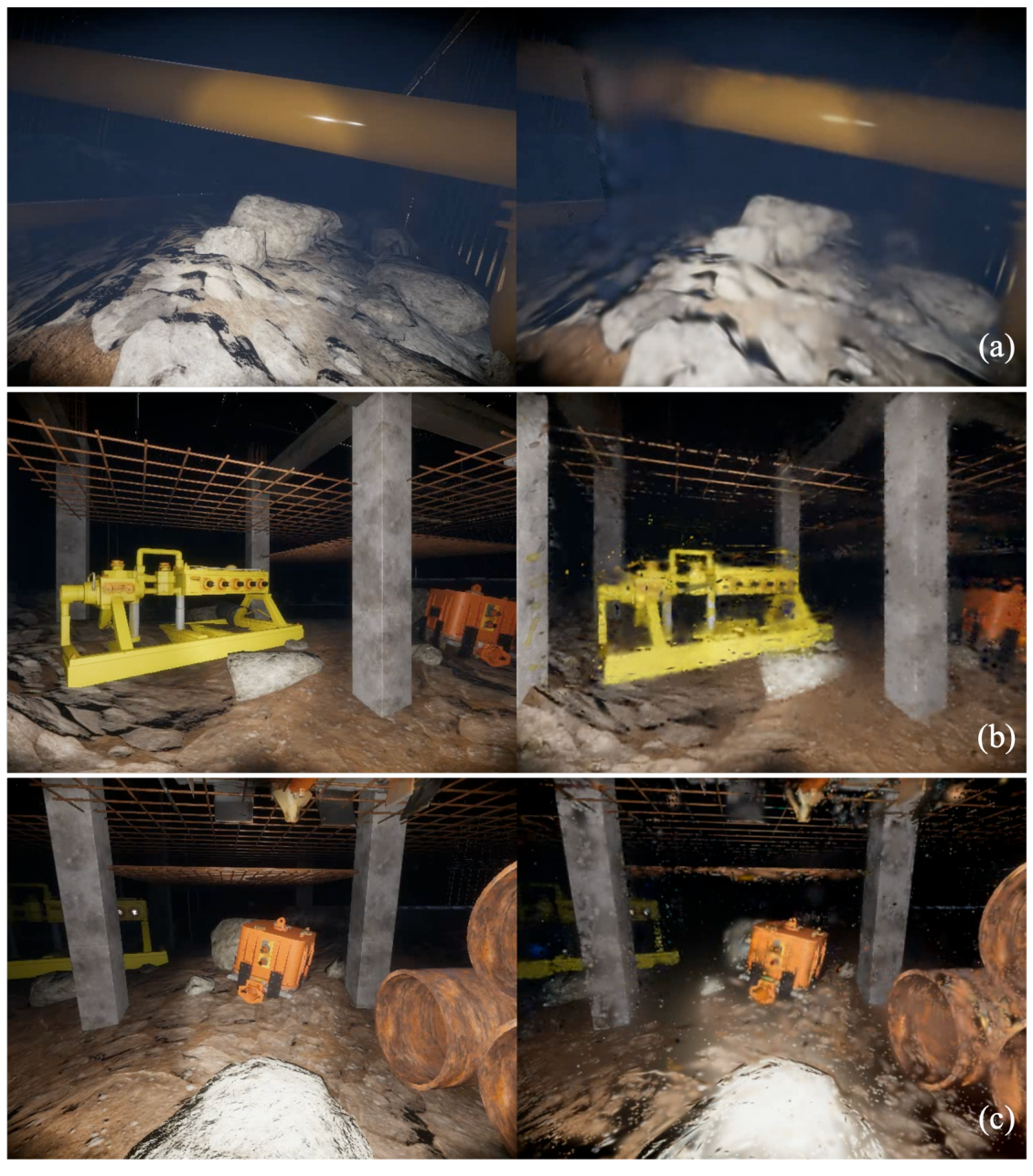

Figure 3. Real-time novel view synthesis with 3D Gaussian Splatting using simulated underwater envirionment.

During active teleoperation, the DAVS and MonoGS modules function synergistically. Once the DAVS optimizer determines the optimal occlusion-free exocentric pose $c^*$, the rendering pipeline directly queries the active MonoGS mapping thread. A rigorous transformation matrix converts the standard ROS body frame convention (X-forward, Y-left, Z-up) into the OpenCV camera convention [17] (X-right, Y-down, Z-forward) required by the rasterizer.

The expected 2D pixel color $C(p)$ for the exocentric view is synthesized through front-to-back alpha compositing:

$$C(p) = \sum_{i \in N} c_i \alpha_i \prod_{j=1}^{i-1} (1 - \alpha_j)$$

where $c_i$ is the SH-evaluated RGB color and $\alpha_i$ is the activated opacity from the live MonoGS map. Finally, to physically anchor the operator's spatial awareness, the ROV's bounding box is projected into this dynamically synthesized image as a golden yellow wireframe overlay utilizing the vehicle's TF pose.

### *D. Vibrotactile Proximity Cues*

To distribute spatial awareness to the somatosensory system, we integrate a torso-mounted vibrotactile suit. While DAVS provides proactive spatial context, the haptic subsystem acts as a reactive safety shield. The ROV collision geometry is discretized into a bounding volume $\mathcal{B}$, partitioned into $N$ distinct sectors $\mathcal{B}_i$ corresponding to the spatial layout of the actuators. For each sector, high-frequency spatial queries compute the minimal Euclidean clearance $d_i(t)$ between the ROV bounding sector and the surrounding facility mesh $\mathcal{E}$ at time $t$:

$$d_i(t) = \min_{p \in \mathcal{B}_i, q \in \mathcal{E}} |p - q|_2$$

To map this physical clearance to vibrotactile intensity $I_i(t)$ without overwhelming the operator in free space, we utilize a bounded, non-linear activation function:

$$I_i(t) = I_{max} \cdot \Phi\big(d_i(t)\big) \cdot \left(\frac{d_{thresh} - d_i(t)}{d_{thresh} - d_{min}}\right)^{\gamma}$$

where $I_{max}$ is the upper saturation limit, $d_{thresh}$ is the tunable safety activation radius, and $d_{min}$ is the critical collision threshold. The exponent $\gamma>1$ shapes the urgency curve, ensuring intensity scales aggressively nly as the vehicle approaches a boundary. The gating function $\Phi(d)$ enforces strict silence outside the threat radius using the Heaviside step function $H(\cdot)$:

$$\Phi(d) = H(d_{thresh} - d) \cdot H(d - d_{min})$$

Because the system evaluates cognitive workload via fNIRS, exact timing consistency between sensory inputs and neurological responses is critical. Synchronization is enforced via message timestamps propagated from ROS to Unity. While DAVS outputs at 20-25 fps depending on scene complexity, Unity buffers and interpolates intermediate frames to maintain a stable 90 Hz display refresh. Every visual viewpoint toggle and haptic activation event generates a time-stamped synchronization tag. Visual-to-haptic latency is strictly bounded below 20 ms to prevent sensorimotor mismatch.

## IV. Experimental Methodology and fNIRS Acquisition

### *A. Study design and hypotheses*

We conducted a controlled, within-subject experiment with 30 participants to evaluate how distributing spatial awareness across distinct sensory channels affects teleoperation performance and cognitive workload under varying network conditions. The study utilized a 3×4 repeated-measures factorial design. The first independent variable was the Assistance Modality, which featured three distinct levels: *1)* Egocentric (Baseline): Standard 2D onboard camera feed. *2)*: Haptic (Reactive): Egocentric view augmented with the torso-mounted vibrotactile proximity cues. *3)*: Exocentric (Proactive): The optimized 3DGS-DAVS third-person viewpoint without haptic feedback. A combined Exocentric + Haptic condition was intentionally excluded from this study. Integrating simultaneous dynamic visual shifting with localized vibrotactile alerts risks cross-channel sensorimotor interference and unpredictable spikes in cognitive load. Isolating the modalities allowed us to cleanly decouple and evaluate the distinct neural correlates of proactive spatial planning versus reactive boundary avoidance. The second independent variable was Communication Delay, simulating the high-latency acoustic links typical of subsea deployments. We tested four artificial latency levels injected into the control and telemetry loop: 0.0 s, 0.2 s, 0.5 s, and 1.0 s.

We test the following directional hypotheses:

H1 (Latency Resilience): Exocentric viewing will preserve global spatial awareness under high communication delays, resulting in lower trajectory deviation and faster completion times compared to Egocentric and Haptic modes.

H2 (Reactive Offloading): Haptic proximity cues will significantly reduce localized collision rates at low-to-moderate delays by providing immediate boundary warnings that bypass visually delayed telemetry.

H3 (Neurophysiological Workload): fNIRS will reveal distinct prefrontal activation ($\Delta HbO$) patterns across modalities. Specifically, the 3DGS-DAVS Exocentric offloading will effectively suppress the sharp spike in cognitive workload that typically occurs in standard Egocentric teleoperation when latency reaches critical thresholds (e.g., ≥0.5 s).

### *B. Task, Environment, and Procedure*

Participants teleoperated the BlueROV2 through a confined, highly simulated flooded industrial facility featuring narrow clearances and complex pipe networks. The task concluded when the ROV successfully navigated to a designated target zone, with individual trials lasting approximately 90-180 seconds to capture a full hemodynamic response profile. To mitigate learning and order effects, all 12 conditions (3 modalities×4 delays) were completed by each participant in a randomized sequence utilizing Latin-square counterbalancing. Prior to the recorded trials, participants completed a 5-minute training phase in a separate test environment to internalize the vehicle dynamics, the 3DGS-DAVS interface, and the haptic mappings. To ensure clean neurophysiological data acquisition, a baseline period was collected at the beginning of each condition while participants fixated on a center marker under constant illumination. Furthermore, a strict 2-minute resting phase was enforced between conditions to allow hemodynamic signals to fully return to baseline.

Vehicle state, control inputs, and haptic activation events were time-stamped and logged at 90 Hz with Unity engine. We then extract the following objective behavioral metrics including: *(1)* Trajectory Deviation (RMSE): The root-mean-square error between the executed path and the optimal safe centerline. *(2)* Collision Rate: The proportion of total traversal time the ROV bounding volume intersected with the facility geometry. *(3)* Completion Time: Total duration from task start to target arrival.

### C. fNIRS Acquisition and Preprocessing

Neurophysiological workload was quantified using an NIRx fNIRS system featuring 16 sources and 15 detectors sampled at 10 Hz [18]. We recorded optical density at wavelengths of 760 nm and 850 nm. Regions of Interest (ROIs) focused on the anterior prefrontal cortex (APFC) and bilateral dorsolateral prefrontal cortex (DLPFC), which are regions highly associated with executive control, spatial working memory, and workload regulation. Raw intensity signals were converted to optical density and screened using the Scalp Coupling Index (SCI) [19]; channels with SCI<0.5 were excluded to remove poor optode coupling. The signals were band-pass filtered using a finite impulse response (FIR) filter between 0.04 Hz and 0.15 Hz to attenuate slow drifts, cardiac pulsation, and respiratory noise.

Optical density changes $\Delta OD^{\lambda}$ at wavelength $\lambda$ were converted to relative concentration changes of oxygenated $\Delta$HbO and deoxygenated $\Delta$HbR hemoglobin utilizing the modified Beer-Lambert Law [20]:

$$\begin{bmatrix} \Delta HbO(t) \\ \Delta HbR(t) \end{bmatrix} = (\mathrm{E} \cdot d \cdot DPF)^{-1} \begin{bmatrix} \Delta OD^{\lambda_1}(t) \\ \Delta OD^{\lambda_2}(t) \end{bmatrix}$$

where E is the specific extinction coefficient matrix for the chromophores, $d$ is the source-detector separation distance, and $DPF$ is the differential pathlength factor accounting for tissue scattering.

### D. General Linear Modeling (GLM) and Statistical Analysis

We adopted an event-related design where condition predictors were formed by convolving trial onset markers with a canonical hemodynamic response function (HRF). For each channel, task-evoked activation was estimated using a General Linear Model (GLM):

$$Y = X\beta + Z\gamma + \epsilon$$

where $Y$ is the observed $\Delta$HbO time series, $X$ is the design matrix containing the HRF-convolved task predictors, $\beta$ is the vector of estimated activation amplitudes (the primary indicator of cognitive workload). To enhance signal robustness, $Z\gamma$ represents nuisance regressors including linear drift terms and short-separation channel signals to regress out superficial systemic physiological components, while $\epsilon$ is the residual error. Statistical inference was performed on the ROI-averaged $\beta$ values per subject. Behavioral metrics and fNIRS ROI outcomes were analyzed for main and interaction effects using a 3×4 repeated-measures ANOVA. Nonparametric alternatives (e.g., Wilcoxon signed-rank with Benjamini-Hochberg correction) were utilized when normality assumptions were violated.

## V. Results

Across all behavioral, subjective, and neurophysiological metrics, pairwise post-hoc comparisons were evaluated using nonparametric tests. All reported post-hoc p-values are Benjamini-Hochberg (BH) adjusted to control the false discovery rate at α=0.05.

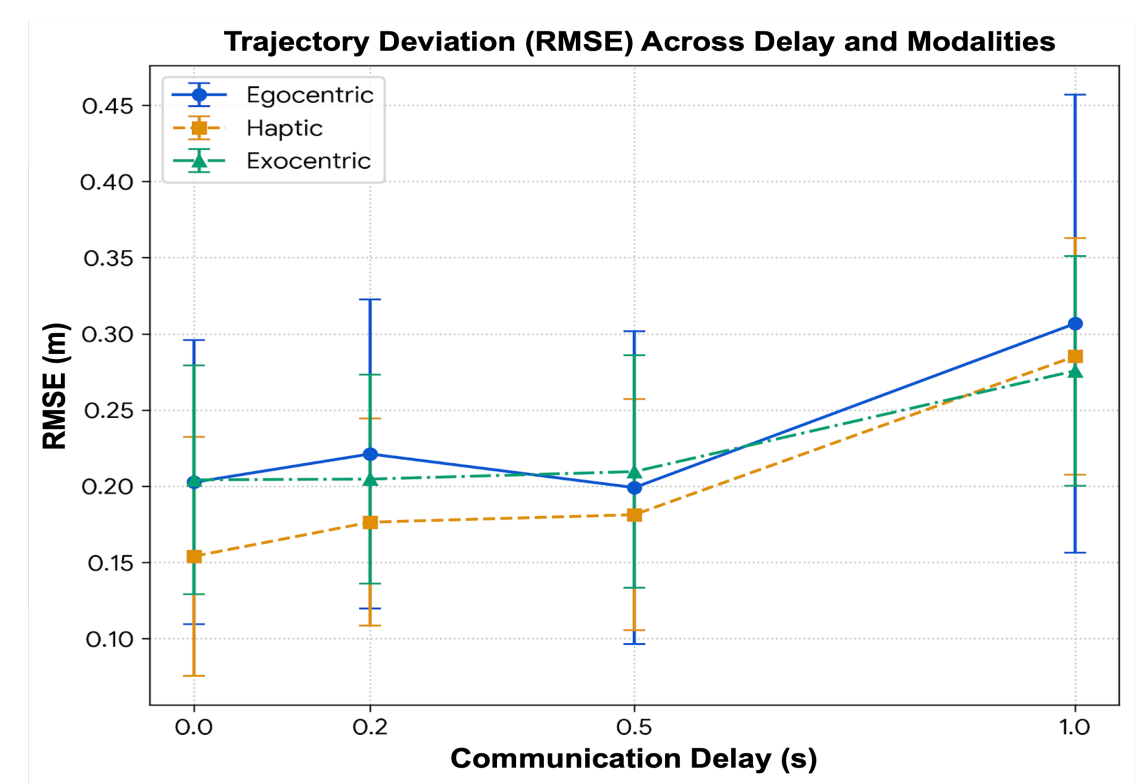

### A. Behavioral Performance Results

We evaluated three behavioral performance metrics, including: *1)*: task completion time, *2)*: average vehicle speed, and *3)*: collision rate across the three feedback/viewing modalities (Egocentric, Haptic, Exocentric) under four communication-delay conditions (0, 0.2, 0.5, and 1.0 s). For each delay level, we conducted within-subject pairwise comparisons using BH-adjusted post-hoc tests (α=0.05) to isolate modality effects while controlling false discoveries.

Completion time and average speed exhibited consistent, delay-dependent trends. At 0.0 and 0.2 s, all modalities were statistically indistinguishable (all $p_adj$>0.05), indicating that under negligible latency, additional feedback channels did not confer measurable efficiency gains. At 0.5 s, Egocentric control showed the first clear degradation, yielding longer completion times (p=0.00365) and reduced speed (p=0.00199) relative to the Haptic condition. At the maximum 1.0 s delay, Exocentric viewing proved the most robust, completing tasks significantly faster than both Egocentric (p=0.00558) and Haptic (p=0.00334) modalities, confirming its capacity to preserve global spatial context under severe latency. Collision rates primarily reflected localized safety. Under moderate delay (0.5 s), the Egocentric condition produced significantly more collisions than the Haptic (p=0.00370) and Exocentric (p=0.0331) conditions. At 1.0 s, Exocentric maintained a lower collision rate than Egocentric (p=0.000105) and Haptic (p=0.00284). This indicates that at extreme latency, proactive 3DGS-DAVS viewpoint assistance preserves collision safety better than reactive haptic alerting alone.

TABLE I. Behavioral Performance Metrics Across Delay and Modality

| Delay | Modality | Completion Time (s) | Average Speed (m/s) | Collision Rate (%) |
|---|---|---|---|---|
| ***0.0 s*** | *Egocentric* | *102.42 ± 3.06* | *0.975 ± 0.029* | *0.022 ± 0.051* |
| | *Haptic* | *102.76 ± 2.25* | *0.979 ± 0.019* | *0.010 ± 0.019* |
| | *Exocentric* | *102.79 ± 3.48* | *0.978 ± 0.026* | *0.004 ± 0.007* |
| ***0.2 s*** | *Egocentric* | *104.09 ± 6.31* | *0.965 ± 0.045* | *0.048 ± 0.085* |
| | *Haptic* | *103.47 ± 4.00* | *0.973 ± 0.032* | *0.034 ± 0.053* |
| | *Exocentric* | *105.24 ± 6.39* | *0.965 ± 0.047* | *0.011 ± 0.019* |
| ***0.5 s*** | *Egocentric* | *107.82 ± 8.88* | *0.965 ± 0.045* | *0.091 ± 0.175* |
| | *Haptic* | *105.08 ± 6.94* | *0.973 ± 0.032* | *0.044 ± 0.097* |
| | *Exocentric* | *106.20 ± 8.94* | *0.965 ± 0.047* | *0.028 ± 0.053* |
| ***1.0 s*** | *Egocentric* | *118.45 ± 16.06* | *0.857 ± 0.078* | *0.123 ± 0.125* |
| | *Haptic* | *118.77 ± 14.53* | *0.913 ± 0.068* | *0.116 ± 0.117* |
| | *Exocentric* | *109.52 ± 10.67* | *0.907 ± 0.083* | *0.052 ± 0.052* |

## B. Trajectory Deviation (RMSE)

To evaluate spatial planning, trajectory deviation was quantified as the root-mean-square error (RMSE) from a consensus tunnel centerline. At low latencies (0.0 s and 0.2 s), the Haptic modality yielded significantly lower trajectory deviation than both Egocentric (p=0.040, p=0.012) and Exocentric (p=0.017, p=0.012) interfaces. This demonstrates that reactive somatosensory feedback effectively enforces strict path adherence when latency is minimal. However, at 0.5 s delay, the main effect of modality vanished (p=0.623), and trajectory deviations homogenized. At 1.0 s delay, trajectory error degraded substantially across all modes (Egocentric: 0.31±0.15 m, Haptic: 0.29±0.08 m, Exocentric: 0.28±0.08 m), with no significant modality differences remaining (p≥0.891). While haptic cues initially improved path adherence, the cognitive burden of severe temporal delay ultimately overwhelmed this advantage.

Figure 4. Trajectory deviation (RMSE) from the consensus centerline across delay levels. Error bars show standard deviation. The Haptic modality significantly improved path adherence at 0.0 s and 0.2 s delays , but performance converged and uniformly degraded across all modalities at 0.5 s and 1.0 s delays.

## C. Neurophysiological Workload (fNIRS Activation)

fNIRS activation was analyzed using GLM-derived beta coefficients averaged within each ROI for $HbO$ (primary) and $HbR$ (confirmatory). An enhanced preprocessing pipeline was utilized, incorporating wavelet-based artifact removal and AR-IRLS robust GLM, coupled with SCI-weighted ROI channel averaging to account for heterogeneous optode coupling quality. Friedman tests and post-hoc Wilcoxon signed-rank tests (BH-FDR corrected) were conducted for each ROI at each delay level.

At 0 s and 0.2 s delay: No significant differences in cortical $HbO$ activation was observed across modalities in any ROI (all *p_adj*>0.05), remaining consistent with the comparable behavioral performance observed at low latency. At 0.5 s delay: Significant prefrontal effects of feedback modality emerged. In the Anterior Prefrontal Cortex (APFC), Friedman testing revealed a significant main effect ($X^2(2)$) =8.54, $p$=0.014, $W$=0.164), with Egocentric showing significantly lower $HbO$ activation than Exocentric ($p$=0.014, $r$=0.540). Furthermore, the Right Dorsolateral Prefrontal Cortex (RDLPFC) displayed significant differences for both Egocentric vs. Exocentric ($p$=0.006, $r$=0.585) and Haptic vs. Exocentric ($p$=0.031, $r$=0.451), with Exocentric producing higher DLPFC activation. These prefrontal effects converge closely with the behavioral data, confirming that 0.5 s serves as the critical delay threshold where Egocentric performance begins to degrade. At 1.0 s delay: No significant prefrontal effects were detected (all *p_adj*>0.05). This suggests that at the highest levels of delay, the overarching cognitive load stemming from latency compensation dominates and effectively washes out any modality-specific prefrontal processing differences. The application of the enhanced pipeline successfully recovered prefrontal effects (APFC, RDLPFC) at the behaviorally meaningful 0.5 s delay, offering direct neurophysiological support for H3.

TABLE II. PREFRONTAL $HbO$ ACTIVATION DIFFERENCES AT CRITICAL DELAY *(0.5 S)*

| Region of Interest (ROI) | Main Effect (Friedman) | Significant Pairwise Contrasts (BH-FDR) | Effect Size (r) |
|---|---|---|---|
| Anterior Prefrontal Cortex (APFC) | *$X^2(2)$ =8.54, p=0.014* | *Egocentric < Exocentric (p = 0.014)* | *0.540 (Large)* |
| Right Dorsolateral Prefrontal Cortex (RDLPFC) | *$X^2(2)$ =4.85, p=0.089* | *Egocentric < Exocentric (p = 0.006)* | *0.585 (Large)* |
| | | *Haptic < Exocentric (p = 0.031)* | *0.451 (Medium)* |

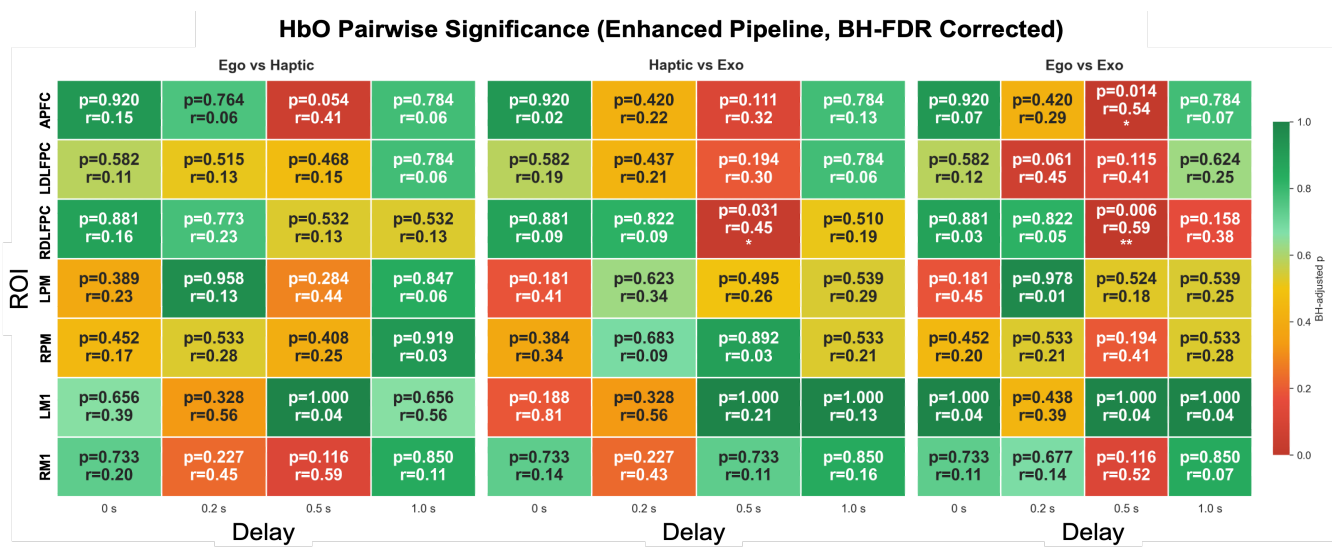

Figure 5. Standalone heatmap showing p-values and effect sizes for all ROI × delay × pairwise comparisons. Warm colors indicate significant effects; key cluster at APFC/RDLPFC × 0.5 s.

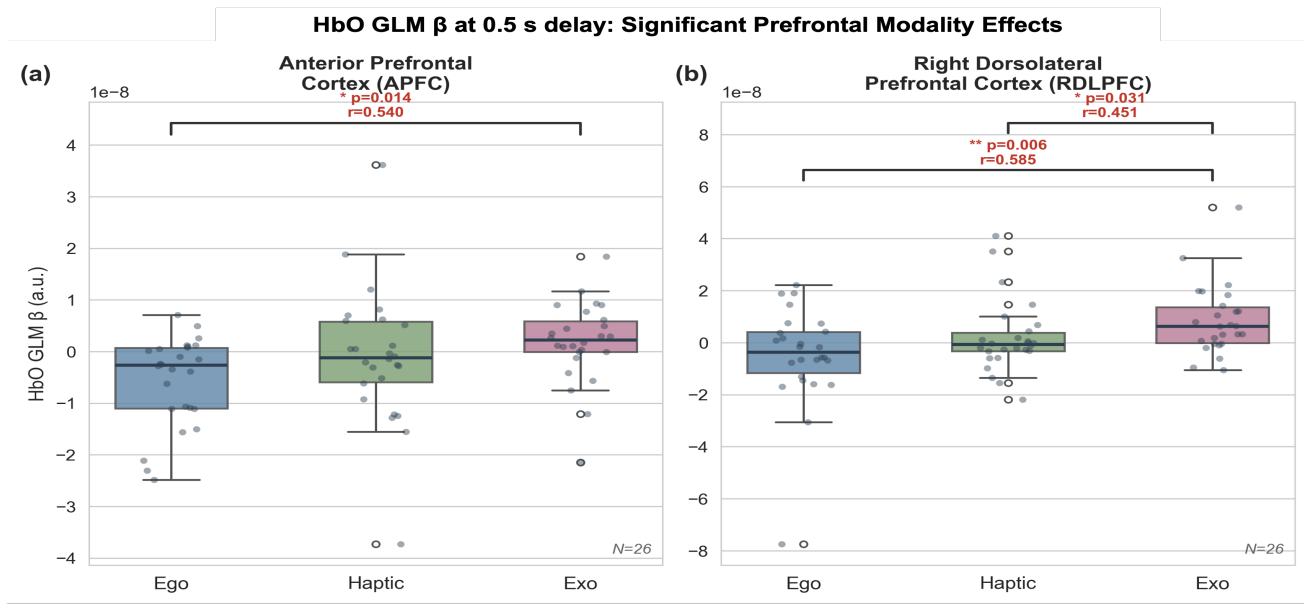

Figure 6. Focused boxplots showing $HbO$ GLM β at 0.5 s delay for the two prefrontal ROIs with significant modality effects. Significance brackets show BH-adjusted p-values and Wilcoxon effect sizes (r).

# VI. DISCUSSION AND LIMITATIONS

The primary objective of this study was to evaluate a novel multimodal architecture designed to resolve the severe perceptual bottlenecks inherent in confined underwater teleoperation. By systematically evaluating Egocentric, Haptic, and Exocentric interfaces across varying network latencies, we sought to determine how computational spatial offloading impacts operator performance and prefrontal workload. The behavioral and neurophysiological results strongly validate the efficacy of this architecture, yielding critical insights for the design of human-in-the-loop robotic systems.

## A. Algorithmic Offloading via Dynamic Viewpoint Optimization

Our behavioral results demonstrate that the efficacy of proactive visual offloading and reactive haptic alerting is highly dependent on communication latency. At minimal delays (0.0 s and 0.2 s), reactive somatosensory feedback effectively enforced strict path adherence, yielding the lowest trajectory deviation. Because the telemetry was nearly instantaneous, operators could seamlessly integrate the localized boundary warnings to optimize their path. However, as latency increased to 0.5 s and 1.0 s, the proactive Exocentric (3DGS-DAVS) interface proved vastly superior. Under a 1.0 s delay, the Exocentric modality completed tasks significantly faster and with fewer collisions than both Egocentric and Haptic modalities. In highly cluttered, low-texture environments where static cameras fail due to structural occlusion, DAVS effectively acts as an algorithmic spatial filter. By dynamically maximizing the Shannon entropy of visible free space, the system successfully buffers the operator against severe temporal delays, proving that stable, global geometric context is more critical than localized alerts during high-latency operations.

### B. Cognitive Disengagement vs. Sustained Executive Control

The most revealing system-level insight stems from the neurophysiological data captured at the critical 0.5 s delay threshold. Contrary to the assumption that lower prefrontal activation universally indicates a superior, lower-workload interface, our fNIRS data captured a classic "cognitive disengagement" phenomenon. At the 0.5 s delay, standard Egocentric teleoperation degraded significantly in both speed and safety. However, this condition elicited the lowest $\Delta HbO$ activation in the Anterior Prefrontal Cortex (APFC). This drop indicates that the perceptual bottleneck became so severe that operators' working memory was overwhelmed; unable to actively maintain a mental 3D map of the delayed environment, their prefrontal resource recruitment collapsed.

Conversely, the Exocentric 3DGS-DAVS modality produced significantly higher $\Delta HbO$ activation in both the APFC and RDLPFC at this same 0.5 s delay, while simultaneously yielding superior behavioral performance and collision avoidance. This proves that the proactive global context provided by the Exocentric view allowed operators to sustain active executive control. Rather than disengaging, their prefrontal cortex remained actively recruited to successfully parse the delayed telemetry and navigate the environment safely. At the extreme 1.0 s delay, these modality-specific prefrontal differences washed out entirely, defining a universal cognitive threshold where the temporal disconnect simply overwhelms human executive function regardless of the interface.

### C. Implications for Subsea Infrastructure Maintenance

The deployment of autonomous systems in complex underwater infrastructure is currently bottlenecked by the brittleness of pure SLAM and autonomy in degraded, low-texture conditions. Consequently, human teleoperation remains mandatory. However, the immense cognitive load of navigating these environments severely limits mission duration and increases the probability of catastrophic hardware loss. The proposed architecture directly addresses deployability. By computationally transferring the burden of 3D spatial updating from the operator to the ROS-Unity backend, this stack preserves operator stamina and accelerates the training pipeline for novice pilots. Furthermore, because DAVS reconstructs the visual scene locally using sparse point clouds, the system is highly resilient to the low-bandwidth, high-latency acoustic communication links typical of subsea deployments.

### D. Limitations and Future Work

While the simulated underwater facility provided a highly controlled environment necessary for clean fNIRS acquisition, it inherently lacks the chaotic hydrodynamic disturbances experienced by physical ROVs. The primary barrier to immediate real-world validation is the logistical difficulty of constructing a physically submerged, highly cluttered infrastructure to safely test severe geometric occlusions. Furthermore, executing concurrent, real-time MonoGS mapping and DAVS optimization imposes immense computational demands. Deploying this dual-pipeline architecture directly on the SWaP-constrained (Size, Weight, and Power) edge hardware of a physical subsea vehicle remains a significant systems-engineering challenge. Consequently, future work will focus on fabricating a scalable, modular underwater testbed and optimizing the MonoGS rasterization pipeline for edge-compute devices to enable rigorous sim-to-real validation using our physical BlueROV2 platform. Additionally, while this system focused entirely on perceptual assistance, future iterations will integrate shared-control autonomy. We aim to explore how low-level autonomous obstacle avoidance policies can be seamlessly blended with our multimodal feedback to further depress prefrontal workload without diminishing the operator's sense of agency.